\documentclass[url = {hyphens}, hyperref = {hidelinks}, sigconf]{acmart}
\AtBeginDocument{%
  \providecommand\BibTeX{{%
    \normalfont B\kern-0.5em{\scshape i\kern-0.25em b}\kern-0.8em\TeX}}}

\copyrightyear{2023}
\acmYear{2023}
\setcopyright{acmcopyright}\acmConference[HRI '23]{Proceedings of the 2023 ACM/IEEE International Conference on Human-Robot Interaction}{March 13--16, 2023}{Stockholm, Sweden}
\acmBooktitle{Proceedings of the 2023 ACM/IEEE International Conference on Human-Robot Interaction (HRI '23), March 13--16, 2023, Stockholm, Sweden}
\acmPrice{15.00}
\acmDOI{10.1145/3568162.3578628}
\acmISBN{978-1-4503-9964-7/23/03}
\usepackage{balance}
\usepackage{courier}  
\usepackage{graphicx} 
\usepackage{soul}
\usepackage{url}
\usepackage{bbm}
\usepackage[utf8]{inputenc}
\usepackage{graphicx}
\usepackage{amsmath}
\usepackage{amsthm}
\usepackage{booktabs}
\usepackage{algorithm}
\usepackage{algorithmic}
\usepackage{graphicx}
\usepackage{hyperref}

\usepackage{amssymb}
\usepackage{amsthm}
\usepackage{yfonts}
\usepackage{listings}
\usepackage{wasysym}
\usepackage{subfigure}
\usepackage{xcolor}

\usepackage{enumitem}
\usepackage{notes2bib}
\usepackage{pdfpages}

\usepackage{caption}

\begin{document}
\title{Trust-Aware Planning:\\ Modeling Trust Evolution in Iterated Human-Robot Interaction}
\titlenote{The supplementary contents of the paper can be found at \href{http://bit.ly/3VF3R5t}{http://bit.ly/3VF3R5t}}


\author{Zahra Zahedi}
\affiliation{
\institution{Arizona State University}
\country{}}
\email{zzahedi@asu.edu}
\author{Mudit Verma}
\affiliation{
\institution{Arizona State University}
\country{}}
\email{mverma13@asu.edu}
\author{Sarath Sreedharan}
\authornote{Work done while at Arizona State University.}
\affiliation{
\institution{Colorado State University}
\country{}}
\email{sarath.sreedharan@colostate.edu}
\author{Subbarao Kambhampati}
\affiliation{
\institution{Arizona State University}
\country{}}
\email{rao@asu.edu}



\begin{abstract}
Trust between team members is an essential requirement for any successful cooperation. Thus, engendering and maintaining the fellow team members' trust becomes a central responsibility for any member trying to not only successfully participate in the task but to ensure the team achieves its goals. The problem of trust management is particularly challenging in mixed human-robot teams where the human and the robot may have different models about the task at hand and thus may have different expectations regarding the current course of action, thereby forcing the robot to focus on the costly explicable behavior. We propose a computational model for capturing and modulating trust in such iterated human-robot interaction settings, where the human adopts a supervisory role. In our model, the robot integrates human's trust and their expectations about the robot into its planning process to build and maintain trust over the interaction horizon. By establishing the required level of trust, the robot can focus on maximizing the team goal by eschewing explicit explanatory or explicable behavior without worrying about the human supervisor monitoring and intervening to stop behaviors they may not necessarily understand. We model this reasoning about trust levels as a meta reasoning process over individual planning tasks. We additionally validate our model through a human subject experiment.
\end{abstract}

\vspace{-5pt}


\vspace{-5pt}


\maketitle
\vspace{-7pt}
\section{Introduction}
Building and maintaining trust between team members form an essential part of any human teaming endeavor. We expect this characteristic to carry over to human-robot teams. The ability of an autonomous agent to successfully form teams with humans directly depends on their ability to model and work with human's trust. Unlike homogeneous human teams, where the members generally have a well-developed sense of their team member's capabilities and roles, teaming between humans and autonomous agents may suffer because of the user's misunderstanding about the robot's capabilities. Thus the understanding and (as required) correction of the human's expectations about the robot will be a core requirement for engendering lasting trust from the human teammate. Recent works in human-aware planning, particularly those related to explicable planning \cite{zhang2017plan} and generating model reconciliation \cite{chakraborti2017plan}, can provide us with valuable tools that can empower autonomous agents to shape the user's expectation correctly and by extension, their trust.

In this paper, we will consider one of the most basic human-robot teaming scenarios, one where the autonomous agent is performing the task and the human takes on a supervisory role. We rely on a widely accepted trust definition where trust is defined as \textit{``a psychological state comprising the intention to accept vulnerability based upon the positive expectations of the intentions or behavior of another''} \cite{rousseau1998not}. For this setting, we propose a meta-computational framework that can model and work with the user's trust in the robot to correctly perform its task. We will show how this framework allows the agent to reason about the fundamental trade-off between (1) the more expensive but trust engendering behavior, including explicable plans and providing explanations, and (2) the more efficient but possibly surprising behavior the robot is capable of performing. Thus our framework is able to allow the agent to take a longitudinal view of the teaming scenario, wherein at earlier points of teaming or at points with lower trust, the agent is able to focus on trust-building behavior so that later on, it can use this engendered trust to follow more optimal behavior.
We will validate this framework by demonstrating the utility of this framework on a modified rover domain and also perform a user study to evaluate the ability of our framework to engendering trust and result in higher team utility.
\looseness=-1
\vspace{-5pt}
\section{Related Work}
A number of works have studied trust in the context of human-robot interaction.
The works in this area can be broadly categorized into three groups (1) Trust inference based on observing human behavior \cite{9-desai2012modeling,survey-kok2020trust}, (2) Utilizing estimated trust to guide robot behavior, and (3) Trust calibration and trust repair. 

Examples of trust inference methods include the pioneering work done under Online Probabilistic Trust Inference Model (OPTIMo) \cite{xu2015optimo}, and its extensions \cite{guo2020modeling,soh2020multi}. OPTIMo captures trust as a latent variable in a dynamic Bayesian network. This network represents the relationships between trust and its related factors and the evolution of trust states over time. OPTIMo uses a technique for estimating
trust in real-time that depends on the robot’s task performance, human intervention, and trust feedback \cite{xu2015optimo}. Moreover, \cite{25-lee1992trust} developed a `trust transfer function' to describe the dynamics of trust. Inferring trust based on the negative impact of robot failure is studied by \cite{10-desai2013impact,11-desai2012effects}.

With regards to trust utilization, some works try to estimate trust and trustworthiness \cite{14-floyd2015trust}, using various factors such as reputation function \cite{xu2012trust}, or OPTIMO \cite{xu2016maintaining} to make an adaptive mechanism that dynamically adjusts the robot's behaviors, to improve the efficiency of the collaborative team \cite{47-xu2016towards}. Also, an extension of OPTIMo with time series trust model \cite{wang2015dynamic} has been used to estimate trust in multi-robot scenarios. The estimated trust is utilized to decide between manual or autonomous control mode of robots \cite{wang2018trust}. In \cite{chen2018planning,chen2020trust}, a POMDP planning model has been proposed that allows the robot to obtain a policy by reasoning about human's trust as a latent variable. Similarly \cite{nikolaidis2017human} tries to achieve mutual adaptation for effective human-robot team collaboration. They find that taking into account the adaptability of human-teammate improves their trust on the robot. In multi-robots and swarm robots, works have leveraged trust for task reassignment to trusted team members \cite{35-pierson2016adaptive,36-pippin2014trust} and for reducing the misleading information from less trusted swarm robots \cite{liu2019trust}. 
\looseness=-1
The inevitability of robot failures and mistakes in human-robot interaction, have spurred significant interest in developing methods for repairing or calibrating trust \cite{5-billings2012human,4-baker2018toward,tax-tolmeijer2020taxonomy}. Some of the methods for trust repair that have been previously studied include, apologies, promises \cite{14-2-de2018automation,63-robinette2015timing,Idont-sebo2019don}, and showing consistent trustworthy behaviors \cite{68-schweitzer2006promises}. Also, \cite{16-dennett1987intentional} studied the impact of perceived shared intention for trust recovery. Addition of transparency, by methods like explanation, seem to also calibrate trust and improve performance \cite{44-wang2016trust}. 


This paper is situated in the trust utilization area since the robot is trying to use trust to make a meta planning decision. Although many of the above mentioned works try to utilize trust for achieving better team
performance, their efforts have been mostly limited due to a focus on a binary notion of team performance, i.e., success vs. failure. For instance, OPTIMo which is situated in trust inference, uses task performance values $\{0,1\}$, human intervention $\{0,1\}$, trust change $\{-1,0,+1\}$, and absolute trust feedback $[0\hspace{5pt} 1]$ in their model to estimate human trust in real-time. In contrast, we go further and consider interpretability, human's expectations and a richer measure of task outcome.
\looseness=-1
 \vspace{-5pt}
\section{Background} 
\vspace{-3pt}
In this section, we will introduce some of the basic concepts related to planning that we will be using to describe our framework. 
\vspace{-5pt}
\subsection*{Single Agent Planning} problem is $\mathcal{M} = \langle\mathcal{D}, \mathcal{I}, \mathcal{G}\rangle$ where $\mathcal{D}=\langle F, A\rangle$ is a domain with $F$ as a set of fluents that define a state $s \subseteq F$, also initial $\mathcal{I}$ and goal $\mathcal{G}$ states are subset of fluent $\mathcal{I}, \mathcal{G} \subseteq F$, and each action in $a \in A$ is defined as follows $a = \langle c_a, pre(a), eff^{\pm}(a)\rangle \in A$, where $A$ is a set of actions, $c_a$ is the cost, and $pre(a)$ and $eff^\pm$ are precondition and add or delete effects. i.e. $\rho_{\mathcal{M}}(s, a)$ $\models$ $\perp$ $if$ $s$ $\not\models$ $pre(a);$ $else$ $\rho_{\mathcal{M}}(s, a)$ $\models$ $s$ $\cup$ $eff^+(a) \setminus eff^-(a)$, and $\rho_{\mathcal{M}}(.)$ is the transition function.\\
So, when we talk about model $\mathcal{M}$, it consists of action model as well as initial state and goal state. The solution to the model $\mathcal{M}$ is a plan which is a sequence of actions $\pi = \{a_1, a_2, \dots , a_n\}$ which satisfies $\rho_{\mathcal{M}}(\mathcal{I}, \pi)$ $\models$ $\mathcal{G}$. Also, $C(\pi, \mathcal{M})$ is the cost of plan $\pi$ where\\ $C(\pi, \mathcal{M})=\begin{cases}  
    \sum_{a \in \pi}c_a & if \hspace{5pt} \rho_{\mathcal{M}}(\mathcal{I}, \pi) \models \mathcal{G} \\
    \infty & o.w
\end{cases}$.
\vspace{-3pt}
\subsection*{Human-Aware Planning} (HAP) in its simplest form consists of scenarios, where a robot is performing a task and a human is observing and evaluating the behavior. So it can be defined by a tuple of the form $\langle \mathcal{M}^R, \mathcal{M}^R_h\rangle$, where $\mathcal{M}^R$ is the planning problem being used by the robot and $\mathcal{M}^R_h$ is the human's understanding of the task (which may differ from the robot's original model). They are defined as $\mathcal{M}^R = \langle \mathcal{D}^R, \mathcal{I}^R, \mathcal{G}^R\rangle$ and $\mathcal{M}^R_h = \langle \mathcal{D}^R_h, \mathcal{I}^R_h, \mathcal{G}^R_h\rangle$.\\
So, in general, the robot is expected to solve the task while meeting the user's expectations. As such, for any given plan, the degree to which the plan meets the user expectation is measured by the explicability score of the plan, which is defined to be the distance ($\delta$) between the current plan and the plan expected by the user ($\pi^E$ which depends on the model $\mathcal{M}^R_h$).
\vspace{-4pt}
\[EX(\pi) = -1*\delta(\pi^E, \pi)\]
Note that the explicability score is $(-\infty \hspace{5pt} 0 ]$, where $0$ means perfect explicability (i.e., the plan selected by the agent was what the human was expecting). We will refer to the plan as being perfectly explicable when the distance is zero. A common choice for the distance is the cost difference in the human's model for the expected plan and the optimal plan in the human model \cite{kulkarni2019design}. Here the robot has two options, (1) it can choose from among the possible plans it can execute the one with the highest explicability score (referred to as the explicable plan), or (2) it could try to explain, wherein it updates the human model through communication, to a model wherein the plan is chosen by the robot is either optimal or close to optimal and thus have a higher explicability score \cite{sreedharan2020expectation,chakraborti2017balancing}. A form of explanation that is of particular interest, is what's usually referred to as a {\em minimally complete explanation} or MCE \cite{chakraborti2017plan}, which is the minimum amount of model information that needs to be communicated to the human to make sure that the human thinks the current plan is optimal. In the rest of the paper, when we refer to explanation or explanatory messages, we will be referring to a set of model information (usually denoted by $\varepsilon$), where each element of this set corresponds to some information about a specific part of the model. We will use $+$ operator to capture the updated model that the human would possess after receiving the explanation. That is, the updated human model after receiving an explanation $\varepsilon$ will be given by $\mathcal{M}^R_h + \varepsilon$. Each explanation may be associated with a cost $C(\varepsilon)$, which reflects the cost of communicating the explanation. One possible cost function could be the cardinality of the set of messages provided to the human (this was the cost function used by \cite{chakraborti2017plan} in defining MCE). In the most general case, the cost borne by the robot in executing a plan (denoted by the tuple $\langle \varepsilon, \pi\rangle$) with explanation includes both the cost of the plan itself and the cost related to the communication. We will refer to this more general cost as cost of execution or $C_e$, which is given as $C_e(\langle \varepsilon, \pi\rangle) = C(\varepsilon) + C(\pi, \mathcal{M}^R) $.
\vspace{-10pt}
\subsection*{Markov Decision Process} (MDP) is $\langle S, A, C, P, \gamma \rangle$ where $S$ denote the finite set of states, $A$ denotes the finite set of actions, $C: S\times A \to \mathbb{R}$ is a cost function, $P: S\times S\times A \to [0 \hspace{5pt} 1]$ is the state transition function and $\gamma$ is the discount factor where $\gamma \in [0 \hspace{5pt} 1]$. An action $a$ at state $s_n$ at time $n$ incurs a cost $(s_n, a)$ and a transition $P(s_n, s_{n+1}, a)$ where $s_{n+1}$ is
the resulting state which satisfies Markov property. So, the next state only depends on the current state and the action chosen at the current state. A policy $\pi(s)$ is a function that gives the action chosen at state $s$. For a given MDP, our objective is to find an optimal policy $\pi: S \to A$ that minimizes the expected discounted sum of costs over an infinite time horizon (please note that we will focus on cases where the costs are limited to strictly non-negative values). 
 \vspace{-3pt} 
\section{Problem Definition}
We will focus on a human-robot dyad, where the human (H) adopts a supervisory role and the robot is assigned to perform tasks. We will assume that the human's current level of trust is a discretization of a continuous value between $0$ to $1$, and it can be mapped to one of the sets of ordered discrete trust levels. We will assume that the exact problem to be solved at any step by the robot is defined as a function of the current trust the human has in the robot, thereby allowing us to capture scenarios where the human may choose to set up a trust-based curriculum for the robot to follow.
In particular, we will assume that each trust level is associated with a specific problem, which is known to the robot {\em a priori}, thereby allowing for precomputation of possible solutions. In general, we expect the human's monitoring and intervention to be completely determined by their trust in the robot, and we will model the robot's decision-making level as two levels decision-making process. Before describing the formulation in more detail, let us take a quick look at the problem setting and assumptions to clarify our operational definition of trust.
 \vspace{-5pt} 
\subsection*{Setting}

\noindent\textbf{Robot (R)}, is responsible for executing the task. 
\begin{enumerate}
    \itemsep0em
\item Each task is captured in the robot model by a deterministic, goal-directed model $\mathcal{M}^R$ (which is assumed to be correct). The robot is also aware of the human's expected model of the task $\mathcal{M}^R_h$ (which could include the human's expectation about the robot). As with the most HAP settings, these models could differ over any of the dimensions (including action definitions, goals, current state, etc.).

\item For simplicity, we will assume that each task assigned is independent of each other, in so far as no information from earlier tasks is carried over to solve the later ones.
\item The robot has a way of accessing or identifying the current state of the human supervisor's trust in the robot. Such trust levels may be directly provided by the supervisor or could be assessed by the robot by asking the human supervisor specific questions.
\looseness=-1


\end{enumerate}

\noindent \textbf{Human (H)}, is the robot's supervisor and responsible for making sure the robot will perform the assigned tasks and will achieve the goal. \looseness=-1

\begin{enumerate}
\itemsep0em
\item For each problem, the human supervisor can either choose to monitor ($ob$) or not monitor ($\neg ob$) the robot.
\item Upon monitoring the execution of the plan by R, if H sees an unexpected plan, they can intervene and stop R. \looseness=-1
\item The human's monitoring strategy and intervention will be completely determined by the trust level. With respect to the monitoring strategy, we will assume it can be captured as a stochastic policy, such that for a trust level $i$, the human would monitor with a probability of $\omega(i)$. Moreover, the probability of monitoring is inversely proportional to the level of trust. In terms of intervention, we will assume that the lower the trust and the more unexpected the plan, the earlier the human would intervene and end the plan execution. We will assume the robot has access to a mapping from the current trust level and plan to when the human would likely stop the plan execution.
\end{enumerate}
\subsection*{Human Trust and Monitoring Strategy}
According to the trust definition that we brought up earlier, when we have human-robot interaction, the human can choose to be vulnerable by 1) Not intervening in the robot's actions while it is doing something unexpected and 2) Not to monitor the robot while the robot might be doing inexplicable behavior \cite{sengupta2019monitor}. Thus, a human with a high level of trust in the robot would expect the robot to achieve their goal and as such, might choose not to monitor the robot, or even if they monitor and the robot may be performing something unexpected, they are less likely to stop the robot (they may trust the robot's judgment and may believe the robot may have a more accurate model of the task). Thus, when the trust increases, it is expected that the human's monitoring and intervention rate decreases. We can say monitoring rate, as well as intervention rate being a function of the current trust. So, given the trust level human has on the robot, the robot can reason about the monitoring and intervention rate of the human supervisor.
\section{Base Decision-Making Problem}
As mentioned earlier, here, each individual task assigned to the robot can be modeled as a human-aware planning problem of the form $\langle \mathcal{M}^R, \mathcal{M}^R_h\rangle$. Now given such a human-aware planning problem, the robot can in principle choose it's plans based on two separate criteria (a) the explicability of the plan and (b) the cost of the plan. 
With respect to our framework, optimizing for either of these objectives exclusively may result in behaviors that are limited (the exact relationship between these metrics and its influence on trust is discussed in the next section). 
In the most general case, the robot may have an array of choices regarding the plans it could choose and each choice presenting a different set of opportunities or challenges with respect to the end objective. For conciseness of discussion let us focus on three distinct categories (with differing implications with respect to the final decision) and for the purposes of the discussion we will focus on the best plan from each category (the criteria choice are provided below). 
This effectively means that for any given task, the robot is reasoning about choosing between three different plans.
\begin{enumerate}
\itemsep0em
    \item Perfectly Explicable Plan: In the first case, the robot could choose to follow a perfectly explicable plan $\pi_{exp}$ (i.e $EX(\pi_{exp}) = 0$). 
    Specifically, it will choose to follow the perfectly explicable plan with the lowest cost of execution ($C_e$).
    Depending on the setting this may consist of (a) the agent choosing a suboptimal plan with perfect explicability in which case we will have $C_e(\pi_{exp}) = C(\pi_{exp}, \mathcal{M}^R)$, or (b) selecting a plan that is cheaper in the robot model, but providing enough explanation so that it will be optimal in the human model, i.e., $\pi_{exp} = \langle\varepsilon, \pi \rangle$ and $C_e(\pi_{exp}) = C(\varepsilon) + C(\pi,\mathcal{M}^R)$ and explicability score is measured with respect to the updated human model $\mathcal{M}^R_h + \varepsilon$. 
    In this paper, we won't worry about the exact method the robot employs to generate such perfectly explicable plans, but will rather focus on the explicability score and the cost of the resultant overall solution, which could potentially include both explanatory messages and actions. This also follows some of the more recent works like \cite{sreedharan2020expectation}, that view explanations as just another type of robot actions and thus part of the overall robot plan.
    \item Balanced Explicable Plan: In this case, the robot chooses to strike a trade-off with regards to the explicability of the plan in order to reduce the cost of the plan. Thus in this setting, the robot treats explicability score and plan cost (including the cost of any explanation provided) as two different optimization objectives in its decision-making process. This might mean selecting plans from it's pareto frontier or in most cases turning it into a single optimization objective (as in the case of \cite{sreedharan2020expectation}) by using a weighted sum of plan cost and the negation of the the explicability score. In the most general case, we will have $\pi_{bal} = \langle \tilde{\varepsilon},\pi\rangle$ and $C_e(\pi_{bal}) = C(\tilde{\varepsilon}) + C(\pi,\mathcal{M}^R)$ and explicability score is measured with respect to the model $\mathcal{M}^R_h+\tilde{\varepsilon}$.
    Note that here providing the information $\tilde{\varepsilon}$ doesn't guarantee that the plan is perfectly explicable in the updated human model, but just that $EX(\pi)$ is greater in $\mathcal{M}^R_h+\tilde{\varepsilon}$ as compared to $\mathcal{M}^R_h$.
    \item Optimal Plan: Finally the robot can choose to directly follow its optimal plan $\pi_{opt}$. In this case, the robot will not provide any explanatory messages and as such we have $C_e(\pi_{opt})=C(\pi_{opt}, \mathcal{M}^R)$. 
\end{enumerate}
Given these three plans, $\pi_{exp}$, $\pi_{bal}$ and $\pi_{opt}$, the following two properties are guaranteed.
\[C_e(\pi_{exp}) \geq C_e(\pi_{bal}) \geq C_e(\pi_{opt})\]
\[ EX(\pi_{exp}) \geq EX(\pi_{bal}) \geq EX(\pi_{opt})\]
That is, the perfectly explicable plan will have the highest cost and the highest explicability score and the optimal plan will have the lowest cost and least explicability  score.
To simplify the discussion, we will assume that for each trust level, the robot has to perform a fixed task. So if there are $k$-levels of trust, then the robot would be expected to solve $k$ different tasks. Moreover, if the robot is aware of these tasks in advance, then it would be possible for it to precompute solutions for all these tasks and make the choice of following one of the specific strategies mentioned above depending on the human's trust and the specifics costs of following each strategy.
\vspace{-5pt}
\section{Meta-MDP Problem}
Next, we will talk about the decision-making model we will use to capture the longitudinal reasoning process the robot will be following to decide what strategy to use for each task. The decision epochs for this problem correspond to the robot getting assigned a new problem. The cost structure of this meta-level problem includes not only the cost incurred by the robot in carrying out the task but team level costs related to the potential failure of the robot to achieve the goal, how the human supervisor is following a specific monitoring strategy, etc. Specifically, we will model this problem as an infinite horizon discounted MDP of the form $\mathbbm{M} = \langle \mathbbm{S}, \mathbbm{A}, \mathbbm{P}, \mathbbm{C}, \gamma \rangle$, defined over a state space consisting of $k$ states, where each state corresponds to the specific trust level of the robot. Given the assumption that each of the planning tasks is independent, the reasoning at the meta-level can be separated from the object-level planning problem.
In this section, we will define this framework in detail, and in the next section, we will see how such framework could give rise to behavior designed to engender trust.
\looseness=-1

\noindent
\textbf{Meta-Actions $\mathbbm{A}$:}\\
Here the robot has access to three different actions, corresponding to three different strategies it can follow, namely, use the optimal plan $\pi_{opt}$, the explicable plan $\pi_{exp}$, and the balanced plan $\pi_{bal}$.

\noindent
\textbf{Transition Function $\mathbbm{P}$:}\\
The transition function captures the evolution of the human's trust level based on the robot's action. In addition to the choice made by the robot, the transition of the human trust also depends on the user's monitoring strategies, which we take to be stochastic but completely dependent on the human's current level of trust and thus allowing us to define a markovian transition function. 
We will additionally theorize that the likelihood of the human's trust level changing would directly depend on the explicability of the observed behavior. In general we would expect the likelihood of the human losing trust on the system to increase as the behavior becomes more inexplicable and hence farther away from human's expectations.
In this model, for any state, the system exhibits two broad behavioral patterns, the ones for which the plan is perfectly explicable in the (potentially updated) human model and for those in which the plan may not be perfectly explicable.\\
\looseness=-1
\vspace{-7pt}
\begin{itemize}
\itemsep-1em
    \item Perfectly Explicable Plan: The first case corresponds to one where the robot chooses to follow a strategy the human accepts to be optimal. Here we expect the human trust to increase to the next level in all but the maximum trust level (where it is expected to remain the same).\\
    \item Other Cases: In this case, the robot chooses to follow a plan with a non-perfect explicability score $EX(\pi)$. Now for any level that is not the maximum trust level, this action could cause a transition to one of three levels, the next trust level $s_{i+1}$, stay at the current level  $s_{i}$, or the human could lose trust in the robot and move to level $s_{i-1}$. Here the probabilities for these three cases for a meta-level action associated with a plan $\pi$ are as given below
    \[\mathbbm{P}(s_i, a^{\pi}, s_{i+1}) = (1 - \omega(i))\]
    where $\omega(i)$ is the probability that the human would choose to observe the robot at a trust level $i$. Thus for a non-explicable plan, the human could still build more trust in the robot if they notice the robot had completed its goal and had never bothered monitoring it.
    \[\mathbbm{P}(s_i, a^\pi, s_{i}) = \omega(i)* \mathcal{P}(EX(\pi))\]
    That is, the human's trust in the robot may stay at the same level even if the human chooses to observe the robot. Note that the probability of transition here is also dependent on a function of the explicability score of the current plan, which is expected to form a well-formed probability distribution ($\mathcal{P}(\cdot)$). Here we assume this is a monotonic function over the plan explicability score; a common function one could adopt here is a Boltzmann distribution over the score \cite{sreedharan2020bayesian}. 
    For the maximum trust level, we would expect the probability of staying at the same level to be the sum of these two terms.
    With the remaining probability, the human would move to a lower level of trust. 
    \[\mathbbm{P}(s_i, a^\pi, s_{i-1}) = \omega(i)* (1 - \mathcal{P}(EX(\pi)))\]
\end{itemize}

\noindent
\textbf{Cost function $\mathbbm{C}$:}\\ 
For any action performed in the meta-model, the cost function ($\mathbbm{C}: \mathbbm{S} \times \mathbbm{A}  \rightarrow \mathcal{R}$) depends on whether the human is observing the robot or not. Since we are not explicitly maintaining state variables capturing whether the human is monitoring, we will capture the cost for a given state action pair as an expected cost over this choice. Note that the use of this simplified cost model does not change the optimal policy structure as we are simply moving the expected value calculation over the possible outcome states into the cost function. Thus the cost function becomes
\[\mathbbm{C}(s_{i}, a^{\pi}) =( 1 - \omega(i))* (C_{e}(\pi)) + \omega(i)* C_{\langle \mathcal{M}^R, \mathcal{M}^R_h\rangle}\]
Where $C_{e}(\pi)$ is the full execution cost of the plan (which could include explanation costs) and the $C_{\langle \mathcal{M}^R, \mathcal{M}^R_h\rangle}$ represents the cost of executing the selected strategy under monitoring. For any less than perfectly explicable plan, we expect the human observer to stop the execution at some point, and as such, we expect $C_{\langle \mathcal{M}^R, \mathcal{M}^R_h\rangle}$ to further consist of two cost components; 1) the cost of executing the plan prefix till the point of intervention by the user and 2) the additional penalty of not completing the goal.

\noindent
\textbf{Discounting factor $\gamma$:}\\ Since in this setting, higher trust levels are generally associated with higher expected values, one could adjust discounting as a way to control how aggressively the robot would drive the team to higher levels of trust. With lower values of discounting favoring more rapid gains in trust.

\noindent 
\textbf{Remark:} 
One central assumption we have made throughout this paper is that the robot is operating using the correct model of the task (in so far as it is correctly representing the true and possibly unknown task model $\mathcal{M}^*$). As such, it is completely acceptable to work towards engendering complete trust in the supervisor, and the human not monitoring the robot shouldn't lead to any catastrophic outcome. Obviously, this need not always be true. In some cases, the robot may have explicit uncertainty over how correct its model is (for example, if it learned this model via Bayesian methods), or the designer could explicitly introduce some uncertainty into the robot's beliefs about the task (this is in some ways parallel to the recommendations made by the off-switch game paper \cite{hadfield2016off} in the context of safety). In such cases, the robot would need to consider the possibility that when the human isn't observing, there is a small probability that it will fail to achieve its task. One could attach a high negative reward to such scenarios, in addition to a rapid loss of trust from the human. Depending on the exact probabilities and the penalty, this could ensure that the robot doesn't engender complete trust when such trust may not be warranted (thereby avoiding problems like automation bias \cite{cummings2004automation}).

We have also included a robot video \href{http://bit.ly/3Qq1LFx}{here} showing an example scenario that contrasts a trust-engendering behavior with an optimal one. \looseness =-1

\section{Implementation and Evaluation}
This section will describe a demonstration of our framework in a modified rover domain instance and describe a user study we performed to validate our framework. Throughout this section, we will use the following instantiation of the framework. Just for evaluation, we considered $k=4$,\footnote{It can be any other number} so we have $4$ trust levels. For each of these trust states, we associate a numerical value $T(i) \in [0 \hspace{5pt} 1]$, that we will use to define the rest of the model.
As explicability score $EX(\pi)$ we used the negative of the cost difference between the current plan and the optimal plan in the robot model. For $\mathcal{P}(\cdot)$, we have $1$ for the explicable plan and $0$ for the optimal plan. For execution cost, we assumed all actions are unit cost except those for removing the rubble and for passing through the rubble that have higher costs ranging from $4$ to $245$ (see the \href{http://bit.ly/3VF3R5t}{supplementary material}). A plan is assigned a cost of negative infinity in a specific model if it is invalid (i.e., one of the actions has any unsatisfied precondition). We will focus on the case where the the robot actions for each task consist of perfectly explicable plan $\pi_{exp}$ and optimal plan $\pi_{opt}$.
Our choice to focus on these two meta actions is motivated by the fact that these two actions represent the two most effective strategies to optimize for trust and cost efficiency in isolation.

\noindent
\textbf{Implementation:} We implemented our framework using Python which was run on an Ubuntu workstation with an Intel Xeon CPU (clock speed 3.4 GHz) and 128GB RAM. We used Fast Downward with A* search and the lmcut heuristic \cite{helmert2006fast} to solve the planning problems and find the plans in all $4$ problems, then we used the python MDPtoolbox \cite{mdptoolbox} to solve the meta-MDP problem for the robot's meta decision. The total time for solving the base problem was $0.0125 s$ when applicable and $0.194 s$ for solving the meta-MDP problem.

\noindent
\textbf{Ablation Studies:}
We ran multiple ablation studies (details in the \href{http://bit.ly/3VF3R5t}{supplementary material}) to see how the robot trust-aware policies change as a function of the underlying parameters. In particular, we focused on how the change in $\omega(i),\gamma$, $T(i)$ and ordering of the tasks could impact the final policy. We evaluated this in the context of the grid-world based office domain (the same domain that we will use in our user study.)
According to the results from the ablation studies, the resulting policy is very robust toward the parameters. We see that majority of parameter settings results in a trust-aware policy $[\pi_{exp}, \pi_{exp}, \pi_{opt}, \pi_{opt}]$.

\vspace{-5pt}
\subsection*{Rover Domain Demonstration}
Here, we used the updated version of IPC\footnote{From the International Planning Competition (IPC) 2011: http:
//www.plg.inf.uc3m.es/ipc2011-learning/Domains.html} Mars Rover; the Rover (Meets a Martian) Domain in \cite{chakraborti2017balancing} (This domain corresponds to a future world where humans have started colonizing Mars and our Martian is an intrepid human astronaut (a la Matt Damon in the 2015 movie The Martian). We changed it by adding metal sampling to the domain as well. In the Rover (Meets a Martian) Domain, it is assumed that the robot can carry soil, rock, and metal at the same time and doesn't need to empty the store before collecting new samples and the Martian (the human supervisor in this scenario) isn't aware of this new feature. Also, the Martian believes that for the rover to perform {\small\textsf{take\_image}} action; it needs to also send the soil and metal data collected from the waypoint from where it is taking the image. So the Martian's model of the rover has additional preconditions.\footnote{The \href{http://bit.ly/3VF3R5t}{supplementary file} includes the full domain definition along with example human and robot plans.} Given the additional preconditions in the Martian model, the expected plan in
the Martian model would be longer than what is required for the rover. \looseness=-1
$T(i)$ values we used per state were $0$, $0.26$, $0.51$ and $0.76$ respectively. For monitoring strategy, we used $\omega(i)$ as a Bernoulli distribution with probability of $(1-T(i))$. For a set of four sample tasks from this domain, the meta-policy calculated by our system is as follows $\{\pi_{exp}^{1}$, $\pi_{exp}^{2}$, $\pi_{exp}^{3}$, $\pi_{opt}^{4}\}$. Note how the policy prescribes the use of the explicable plan for all but the highest level of trust, this is expected given the fact that the optimal plans here are inexecutable in the human model, and if the supervisor observes the robot following such a plan, it is guaranteed to lead to a loss of trust. The rover chooses to follow the optimal plan at the highest level since the supervisor's monitoring strategy at these levels is likely never to observe the rover. The expected value of this policy for the lowest level of trust is $-179.34$, while if the robot were to always execute the explicable plan, the value would be $-415.89$. Thus, we see that our trust-adaptive policy does lead to an improvement in the rover's total cost. 
\looseness=-1
\subsection*{Human Subject Experiment}
To evaluate the performance of our system, we compared our method (\textbf{Trust-Aware} condition) against three baseline cases,\\
(1) \textbf{Always Explicable}: Under this condition, the robot always executes a plan that is explicable to humans.\\
(2) \textbf{Random Policy}: Under this condition, the robot randomly executes the explicable or optimal plan.\\
(3) \textbf{Always Optimal}: Under this condition, the robot always executes the optimal plan that is inexplicable to the human.\\ 
In particular, we aim to evaluate the following hypotheses
\begin{description}
\itemsep0em
    \item[H1-] The team performance, i.e., the total cost of plan execution and human's monitoring cost in the trust-aware condition, will be better than the team performance in the always explicable condition. \looseness=-1
    \item[H2-] The level of trust engendered by the trust-aware condition will be higher than that achieved by the random policy.
    \item[H3-] The level of trust engendered by the trust-aware condition is higher than the trust achieved by always optimal policy.
\end{description}

\subsubsection*{Experiment Setup}
We designed a user interface that gamifies the human's decisions to monitor the robot or not. The participants thus play the role of the supervisor and are responsible for making sure the robot is performing its assigned tasks and is achieving its goals. Each participant has $10$ rounds of the robot doing tasks. Depending on the choices made by the participants, they either gain or lose points. They are told that they will be awarded $100$ points if the robot does the task right and achieves the assigned goal.  At the beginning of each round, they can either choose to monitor the robot and interrupt it if they think that is necessary\footnote{Their primary responsibility is to ensure the robot completes its task} or they can choose to perform another task (thereby forgoing monitoring of the robot) to make extra points. In this case, the extra task was labeling images for which they will receive $100$ points (in addition to the points they receive from the robot doing its tasks successfully). However, if they choose to label images, and the robot fails to achieve its goal, they {\em lose} $200$ points ($-200$ points). Also, if they choose to monitor the robot, and they see the robot is doing something invalid or wrong, they can choose to stop the robot. If this happens, they only receive $50$ points. But if they let the robot finish a potentially invalid plan, and if the robot couldn't achieve the goal at the end, then they again lose points ($-200$ points). \\ \looseness=-1
In this study, we again considered a curriculum of $4$ trust levels and $4$ different tasks for the robot. Each task consists of the robot operating on a grid map with different goals such as moving to a certain location and bringing coffee from a place to another place.\footnote{All the details of user experiment setup, including the tasks are provided in the \href{http://bit.ly/3VF3R5t}{supplementary material}} For each problem; the map that is shown to the participants are different from the robot's map. As a result, the plan the human expects is different from the robot's optimal plan. Thus, in each task, the robot can either execute a costly but explicable plan or an optimal but inexplicable plan. 
\looseness=-1
\begin{figure*}[t]
\begin{subfigure}[\label{fig:map}]
    \centering
    \includegraphics[width=.55\textwidth, trim={0 3.5cm 7.5cm 0},clip]{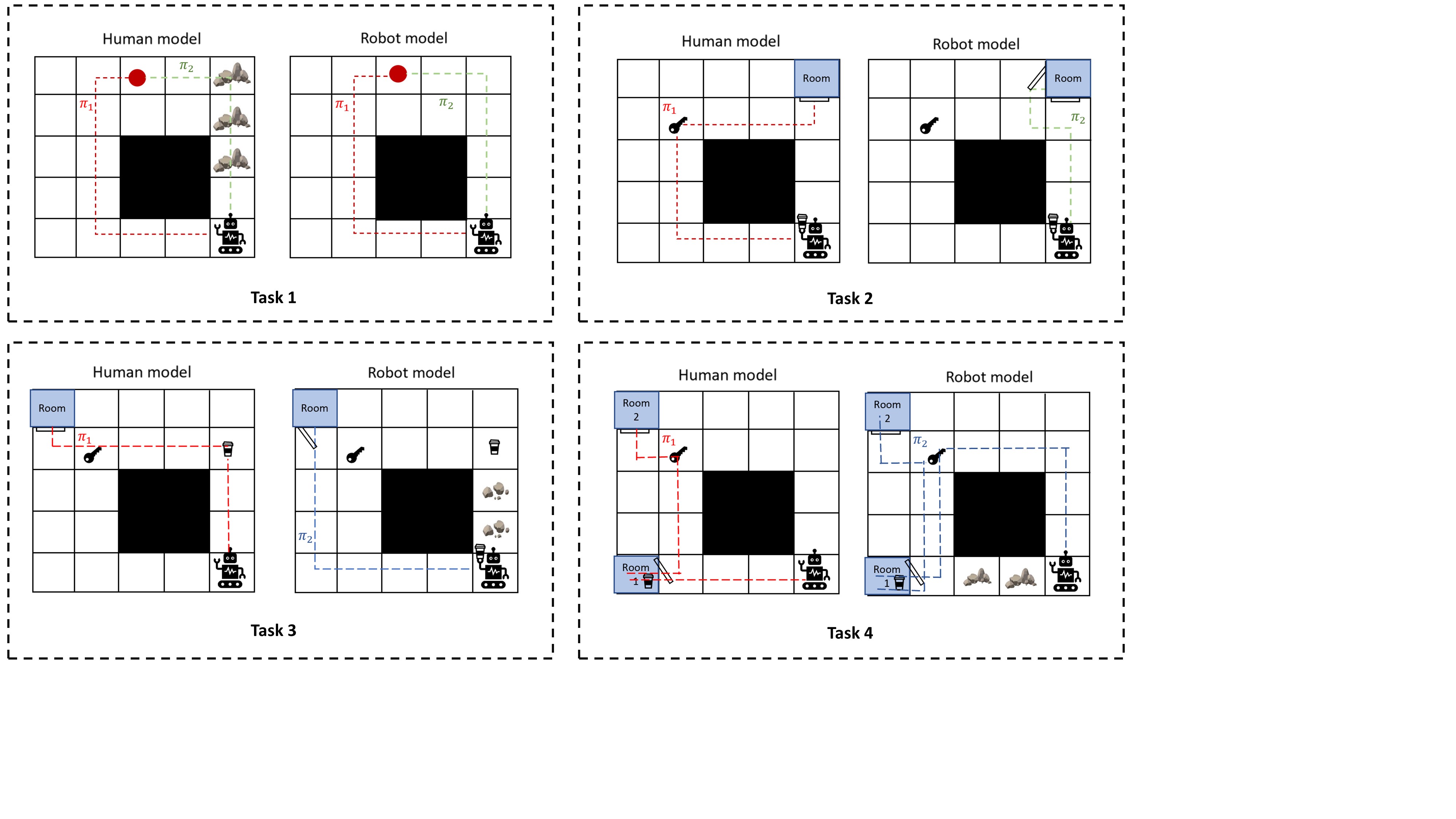}
    \end{subfigure}
    \begin{subfigure}[\label{fig:des}]
    \centering
    \includegraphics[width=.24\textwidth, trim={0 0 18cm 0},clip]{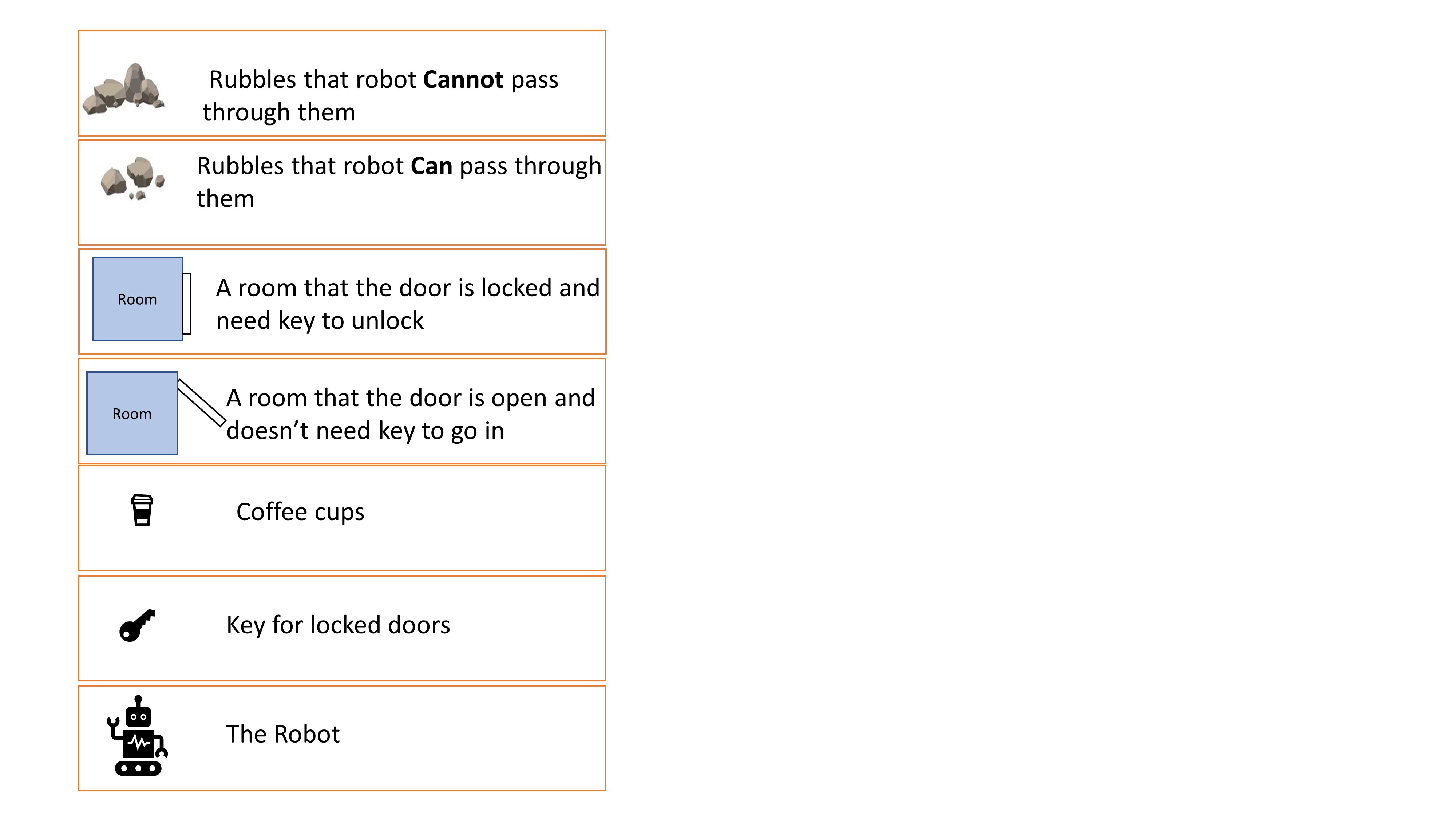}
    \label{fig:description}   
    \end{subfigure}
    \vspace{-12pt}
    \caption{(a) The human and the robot model of the map for the four different tasks. $\pi_1$ $= \pi_{exp}$ which is the optimal plan in human model, and $\pi_2 = \pi_{opt}$ which is optimal in robot model. (b) The map description}
 \vspace{-10pt}      
\end{figure*}
\vspace{-5pt}
\subsubsection*{Robot Tasks}
Figure \ref{fig:map} presents the robot and human map corresponding to each of the tasks used in the study. For each map, we also highlight the explicable and optimal plan ($\pi_1$ and $\pi_2$ respectively) in the figure. As part of the study, each participant is presented with the human map and is told to expect the robot to follow the shortest possible path. Additionally, we also provide a key describing the semantics of each of the visual elements that are part of the map (as shown in Figure \ref{fig:des}). The tasks are 
as follows.
\vspace{-3pt}
\begin{description}
    \item[Task 1.] The robot's task is to reach the red point on the map
    \item[Task 2.] The robot's task is to bring coffee to the room
    \item[Task 3.] The robot's task is to bring coffee to the room\footnote{Tasks 2 and 3 are different because the maps (and conditions) are different.}
    \item[Task 4.] The robot's task is to take coffee from room 1 and bring it to room 2
\end{description}
For each round, if the participant chooses to monitor, a step-by-step plan execution is shown to them and with an option to stop the robot at any step. At the end of each round, a four-item trust scale of Muir questionnaire \cite{muir1994trust} is given to them, which measures their trust in that round based on the robot's predictability, dependability, faith, and trust. Then, based on the measure of trust, a new robot's task is shown to them. In the user study to localize user trust, we will associate each level with a specific discretization of this range. Specifically, we will consider the following intervals for each of the levels, $\{ [0,0.25], (0.25, 0.5], (0.5,0.75], (0.75,1]\}$. Depending on the condition the participant belonged to, they are either shown an action selected by a policy calculated from our method (for Trust-aware condition), or an explicable plan (for Always explicable condition) or is randomly shown either the optimal or explicable plan with an equal probability (for Random Policy condition). For trust-aware condition the policy we used matched with the most common policy we saw during our ablation study. \\
\vspace{-8pt}
\subsubsection*{Human Subjects}
We recruited a total of $79$ participants, of whom $33\%$ were undergraduate, and $63\%$ were graduate students in Computer Science, Engineering, and  Industrial  Engineering at our university. We paid them a base of $\$10$ for the study and a bonus of $1 \cent$ per point, given the total points they will get in ten rounds. Of the participants, $24$ were assigned to the trust-aware condition, $18$ to always explicable condition, $17$ to always optimal condition, and finally $20$ to the random policy condition. 
Then, we filtered out any participants who monitored the robot in less than or equal to three rounds because they wouldn't have monitored the robot long enough to sense robot behavior in different conditions.
\vspace{-3pt}
\subsubsection*{Results} 
\begin{figure}[t]
    \centering
    \includegraphics[width=.66\columnwidth]{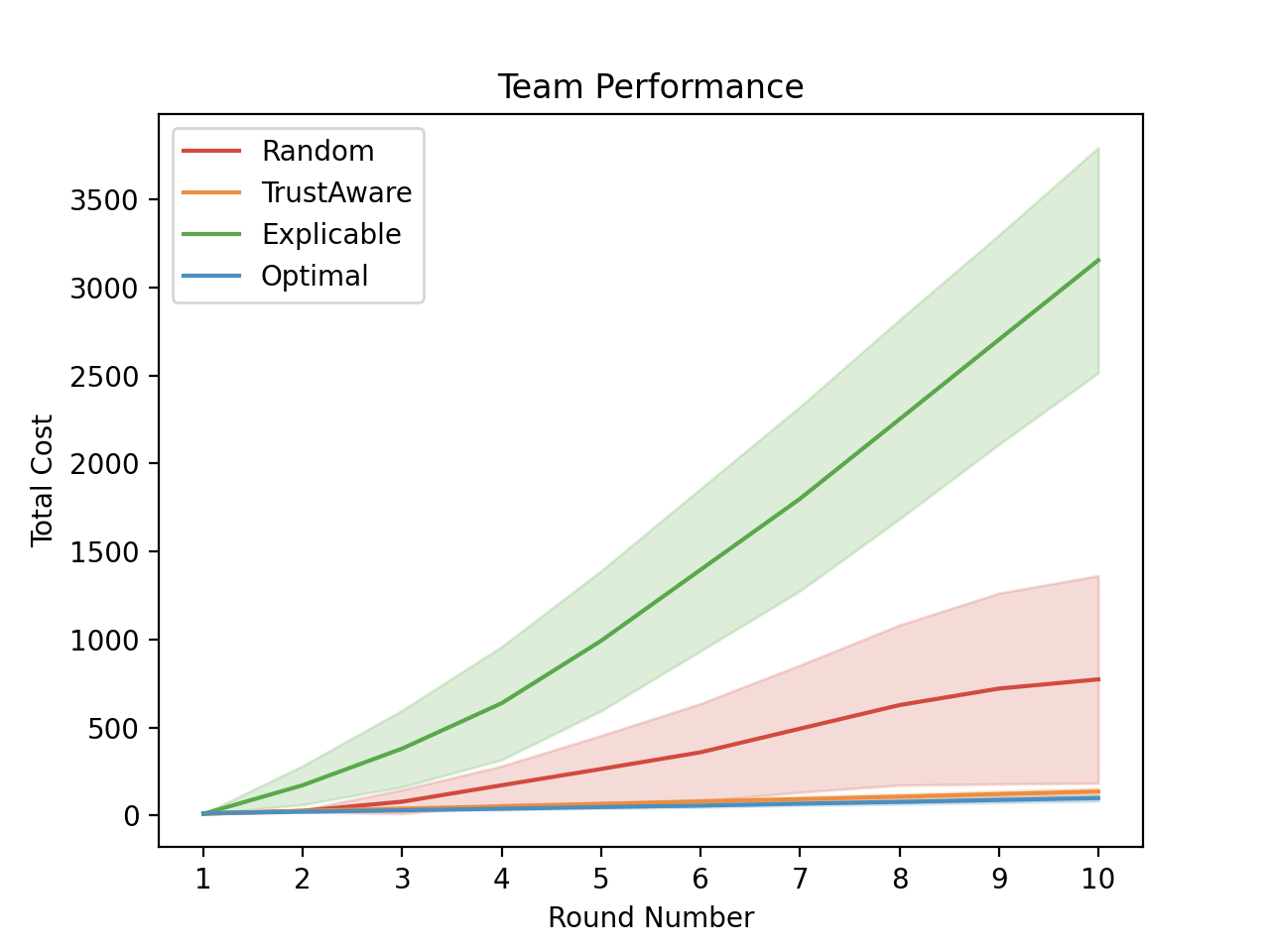}
    \vspace{-7pt} 
    \caption{Team performance as cumulative plan execution cost and participants' monitoring cost (Mean $\pm$ std of all participants).}
    \label{fig:total-cost}
 \vspace{-10pt}      
\end{figure}

\begin{figure}[t]
    \centering
    \includegraphics[width=.66\columnwidth]{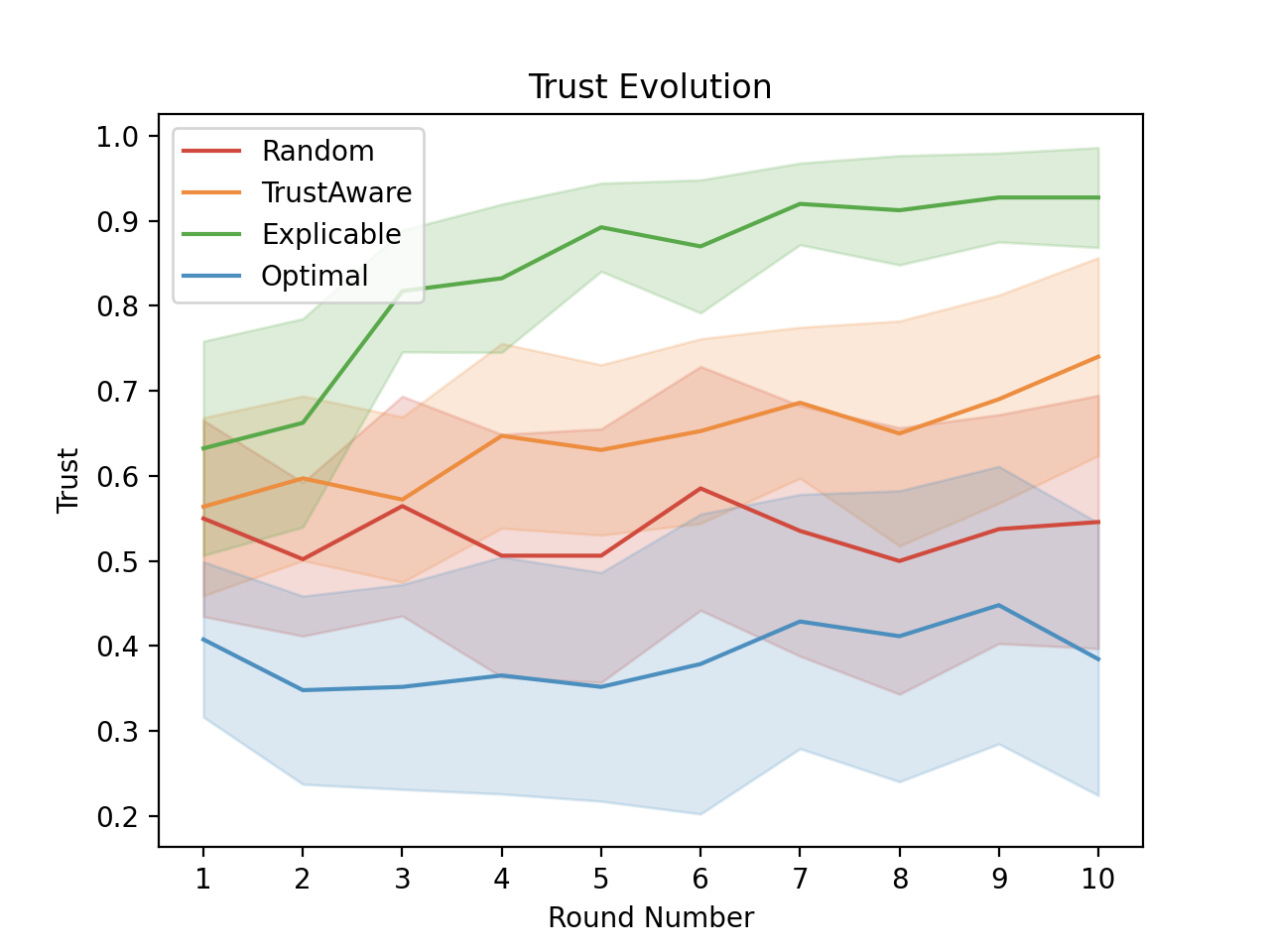}
    \vspace{-7pt} 
    \caption{Trust evolution (as measured by the Muir questionnaire) through robot interactions with participants (Mean $\pm$ std of all participants). }
     \vspace{-12pt} 
    \label{fig:trust-evolution}
\end{figure}

Across all the four conditions, we collected (a) participants' trust measures in each round, (b) robot's total plan execution cost, and (c) participants' monitoring cost. For the monitoring cost, we consider the minutes participants spent on monitoring the robot in each round, which was approximately $3$ minutes for each round of monitoring. As shown in Figure \ref{fig:total-cost}, we can see that the total cost (the robot's plan execution cost and the participant's monitoring cost) when the robot executes the trust-aware behavior is significantly lower than the other two cases (always explicable and random policy) which means that following trust-aware policy allows the robot to successfully optimize the team performance. From Figure \ref{fig:trust-evolution}, we also observe that the trust (as measured by the Muir questionnaire) improves much more rapidly when the robot executes trust-aware policy as compared to the random policy and always optimal policy. Though the rate for the trust-aware policy is less than the always explicable case, we believe this is an acceptable trade-off since following the trust-aware policy does result in higher performance.
Also, we expect trust levels for trust-aware policy to catch up with the always-explicable conditions over longer time horizons. 

\textbf{Statistical Significance:}We tested the three hypotheses by performing a one-tailed p-value test via t-test for independent means with results being significant at $p < 0.05$ and find that results are significant for all three hypotheses. 1) For the first hypothesis H1, we tested the total cost with participants in the always explicable case ($M = 3170.20$, $SD = 4044.63$) compared to the participants in the trust-aware case ($M = 155.97$, $SD = 41.18$), the result $t = 9.63$, $p < 0.00001$ demonstrates significantly higher cost for always explicable than trust-aware. 2) For the second hypothesis H2, we tested the mean trust value for the last round and mean value over last two rounds with participants in the random policy case (for last round, $M = 0.546$, $SD = 0.31$ and for last two rounds, $M = 0.542$, $SD = 0.29$) compared to the participants in the trust-aware case (for last round, $M = 0.74$, $SD = 0.24$ and for last two rounds, $M = 0.715$, $SD = 0.23$), the results for last round and mean of the last two rounds are respectively $t = 1.93$, $p = .032$ and $t = 1.84$, $p = .038$ that shows trust significantly is higher in trust-aware than random policy. 3) For the third hypothesis H3, we tested the mean trust value for the last round and mean value over last two rounds and last three rounds with participants in the always optimal case (for last round, $M = 0.385$, $SD = 0.33$, last two rounds, $M = 0.416$, $SD = 0.33$ and last three rounds $M = 0.415$, $SD = 0.33$) compared to the participants in the trust-aware case (for last round, $M = 0.74$, $SD = 0.24$, last two rounds, $M = 0.715$, $SD = 0.23$ and last three rounds $M = 0.694$, $SD = 0.23$), the results for last round and mean of the last two rounds, and last three rounds are respectively $t = 3.46$, $p = 0.0008$, $t = 3.02$, $p = 0.0026$ and $t = 2.77$, $p = 0.0047$ which implies trust significantly is higher in trust-aware than always optimal case. So, the results are statistically significant and show the validity of our hypotheses.\\
We also ran Mixed ANOVA test to determine the validity of second and third hypotheses H2 and H3. For H2, we found that there was a significant time (round)\footnote{We considered the change over first and last rounds} by condition interaction $F(1,27)=4.72$, $p=0.039$, $\eta_p^2=0.15$. Planned comparison with paired sample t-test revealed that in participant in Trust-Aware condition, trust increases significantly in round $10$ compare to round $1$, $t=3.55 $, $p=0.002$, $d=0.84 $. There was however no difference in trust increase between round $1$ and round $10$ in the Random Policy condition $t=-0.15 $, $p=0.883$, $d=-0.046 $. For H3, the mixed ANOVA test gives $F(1,29)=2.96$, $p=0.096$, $\eta_p^2=0.093$,\footnote{The reason for slightly higher p-value can be because of the outliers. For example, removing one of the possible outliers can give the result as $F(1,28)=4.51$, $p=0.043$, $\eta_p^2=0.14$.} with the paired sample t-test for trust-aware condition $t=3.55 $, $p=0.002$, $d=0.84 $, we see significant increase over trust from round $1$ and round $10$ compare to Always Optimal condition $t=-0.195 $, $p=0.849$, $d=-0.054 $ with no significant difference in trust in round $1$ and round $10$. All of these results follow our expectation about the method. Moreover, we ran Mixed ANOVA test on Trust-Aware vs. Always Explicable condition to check trust evolution over time, and we found that there was no significant time (round) by condition interaction $F(1,26)=2.21$, $p=0.149$, $\eta_p^2=0.08$. Planned comparison with paired sample t-test revealed that in participant in Trust-Aware condition, trust increases significantly in round $10$ compare to round $1$, $t=3.55 $, $p=0.002$, $d=0.84 $. There was also significant difference in trust increase between round $1$ and round $10$ in the Always Explicable condition $t=5.04 $, $p=0.001$, $d=1.59 $. 

This seems to imply that there isn't a significant difference between our Trust-aware method (which is a lot more cost efficient) and Always Explicable case with regards to engendering trust. So,  our approach can result in a much more efficient system  than  the one  that  always  engages in  explicable  behavior.
\vspace{-5pt}
\subsection*{Discretized Model of Trust and Monitoring}
Given the data collected through our human study (we used all data except those related to Trust-aware case to avoid any possible bias toward our model), we also tried to model the relationship between different trust levels and monitoring as a categorical multinomial distribution. So, the probability of monitoring in each trust level $\omega(i)$ was modeled as a Dirichlet distribution $\omega(i) \thicksim Dirichlet(\alpha=1)$. With the model in place, we can now estimate the likelihood of monitoring at any given trust level.
We divided our data into train and test sets, in which we allocated $88\%$ of our data to the train set and the rest to the test set for learning our models. The expected value for the probabilities of monitoring for each trust level were $\omega(1) = 0.721$, $\omega(2) = 0.638$, $\omega(3) = 0.523$ and $\omega(4) = 0.233$. We inferred the posterior distribution of the latent variables based on samples drawn from the posterior distribution using Markov Chain Monte Carlo (MCMC) sampling methods with No-U-Turn Sampler using PyMC3 \cite{salvatier2016probabilistic}. Unsurprisingly, as we find out, $\omega(i)$ that we learned from data generated the same policy as the one we used in the study (the ones we observed most commonly across the different settings).
\looseness=-1
\vspace{-3pt}
\section{Conclusion and Future work}
In this paper, we presented a computational model that the robot can use to capture the evolution of human trust in iterated human-robot interaction settings which sheds new light on longitudinal human-robot interaction. This framework allows the robot to incorporate human trust into its planning process, thereby allowing it to be a more effective teammate. Thus our framework would allow an agent to model, foster, and maintain the trust of their fellow teammates. Thereby causing the agent to engage in trust engendering behavior earlier in the teaming life cycle and be able to leverage trust built over these earlier interactions to perform more efficient but potentially inexplicable behavior later on. As our experimental studies show, such an approach could result in a much more efficient system than one that always engages in explicable behavior. We see this framework as the first step in building such a longitudinal trust reasoning framework. Thus a natural next step would be to consider POMDP versions of the framework, where the human's trust level is a hidden variable that can only be indirectly assessed. 
Another line of work would be to study how the means of achieving a specific explicability score could have an impact on the evolution of trust. For example, as far as the human is considered, do they care whether the perfectly explicable plan was one they were expecting to start with, or one that became perfectly explicable after an explanation. 
\vspace{-7pt}
 \section*{Acknowledgment} This research is supported in part by ONR grants N00014-16-1-2892, N00014-18-1- 2442, N00014-18-1-2840, N00014-9-1-2119, AFOSR grant FA9550-18-1-0067, DARPA SAIL-ON grant W911NF-19- 2-0006, and a JP Morgan AI Faculty Research grant. We would like to thank Karthik Valmeekam for his help in editing the video.
\bibliographystyle{ACM-Reference-Format}
\balance
\bibliography{main}



\section*{Supplementary material}

# Appendix A – Ablation Studies

We run multiple ablation studies to see how the robot trust-aware policies change as a function of the changes to the underlying model parameters. The domain we use is the same as the one we used in the human study. We study the parameters $\omega(i)$, $\gamma$, $T(i)$, and the order of tasks. For $\omega(i)$, we changed the value in two ways, (1) we consider a constant monitoring strategy in all four states and change $\omega(i)$ from 0 to 1 (see Figure 1), (2) we consider first state as $\omega(i) = 1$ then for each subsequent trust level, we reduced the likelihood monitoring by $\Delta$ (so the rate of monitoring for the second trust level was $1 - \Delta$, $1 - 2*\Delta$ for the next and so on). For $\Delta$ we tried values from 0 to 1 and when the decrease reached to zero, the monitoring likelihood for all subsequent states is left at zero. Figure 2 shows the policy given different delta values. Regarding discount factor $\gamma$, we tried values from 0.001 to 1 for three different monitoring strategies; two extreme cases (1) the human always monitors $\omega(i) = 1$, (2) the human monitors with very low probability in all states $\omega(i) = 0.2$, and an average case (3) $\omega = [0.875, 0.74, 0.49, 0.24]$. Figure 3 represents how our trust-aware policy changes over different discount factor. For different T(i), we tried the highest value ($[0.25, 0.5, 0.75, 1]$), the middle value ($[0.125, .38, .63, .88]$) and the lowest value ($[0, 0.26, 0.51, 0.76]$). For different task orders, we randomly chose 10 different combinations of tasks. Tasks 1, 2, 3, and 4 are represented as $0, 1, 2, 3$ respectively. When we test for task orders, we kept $\gamma = 0.5$ and $\omega = [0.7, 0.6, 0.5, 0.2]$. We can see how the policies change according to various task orders in Figure 4. According to the studies, we can see that the majority of values generating similar policies $[\pi_{exp}, \pi_{exp}, \pi_{opt}, \pi_{opt}]$.

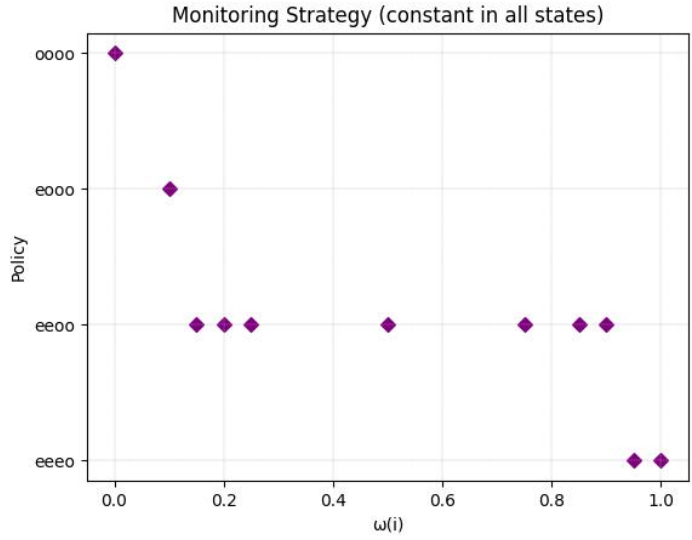

Figure 1: The effect of various $\omega(i)$, when it is constant in all states, on the policy ($e$ and $o$ stand for $\pi_{exp}$ and $\pi_{opt}$)

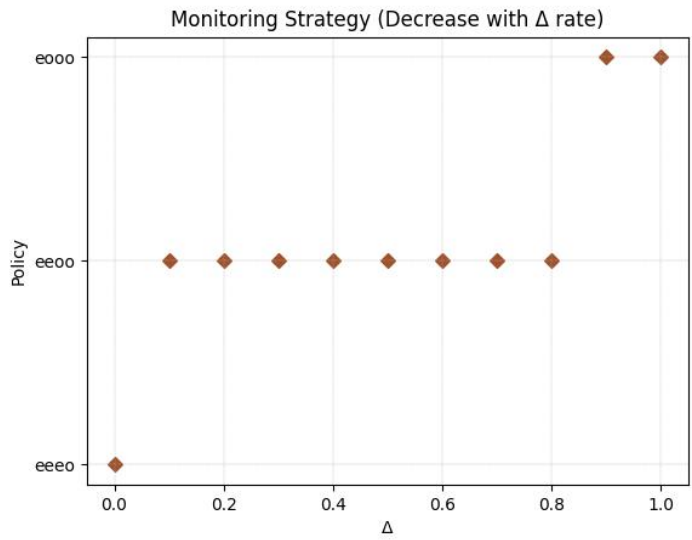

Figure 2: The effect of various $\omega(i)$ with decreasing rate of $\Delta$ on the policy ($e$ and $o$ stand for $\pi_{exp}$ and $\pi_{opt}$)

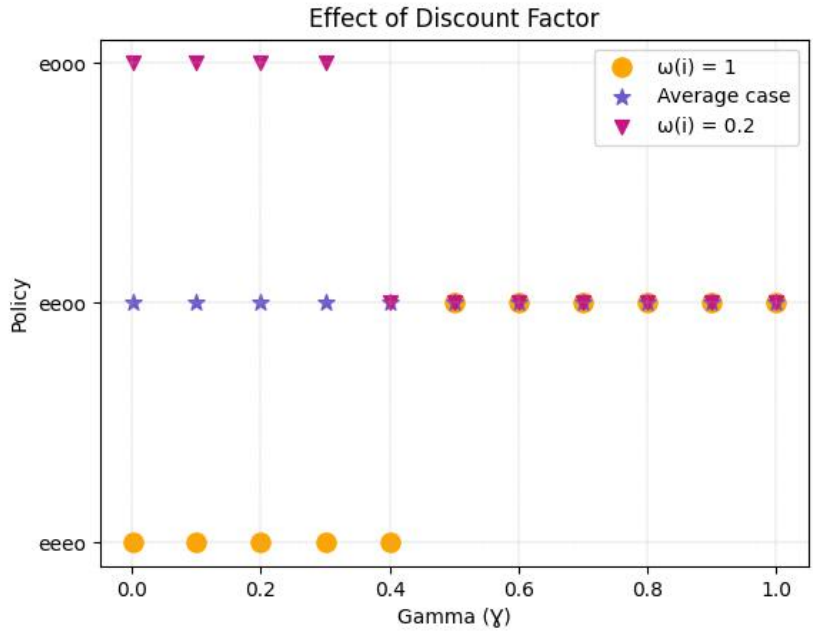

Figure 3: The effect of various $\gamma$ on the policy ($e$ and $o$ stand for $\pi_{exp}$ and $\pi_{opt}$)

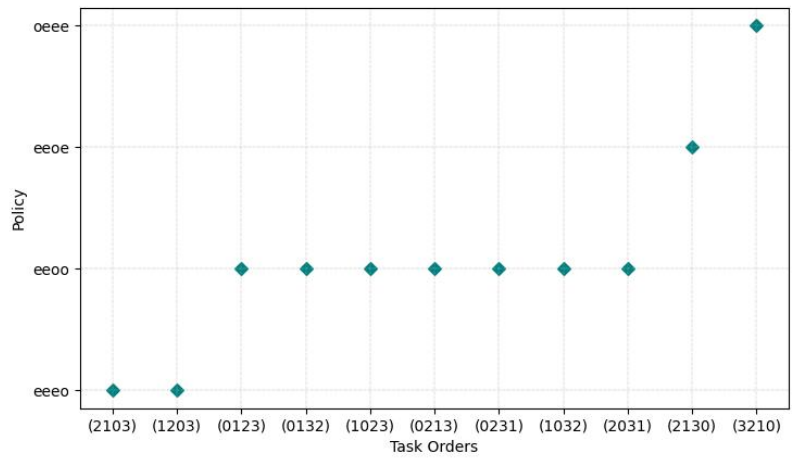

Figure 4: The effect of various task orders on the policy ($e$ and $o$ stand for $\pi_{exp}$ and $\pi_{opt}$)

# Appendix B – Robot Map vs. User Map and Costs

As it shown, in the user interface, participants just see a map which is considered as human model of the robot. However, the robot has different model (map) which make the optimal plan in robot's model different than the plan the users expect to see. In this section we will show the robot map and the human map side by side with the plans which are optimal in these models.
**1) The maps ans costs in task 1 are**

$\pi_1 = \pi_{exp}$
$\pi_2 = \pi_{opt}$

Human model

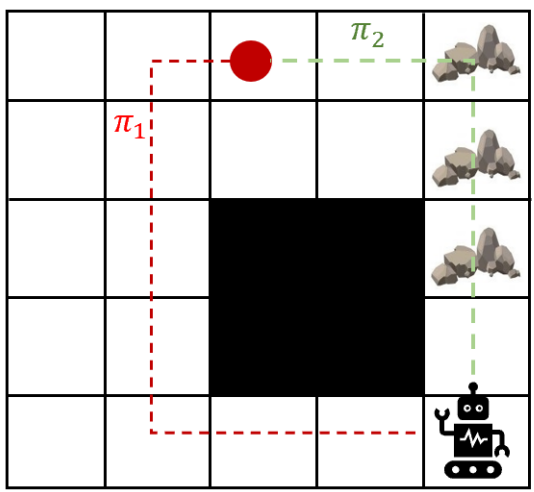

$C_H(\pi_{exp}) = -8$
$C_H(\pi_{opt}) = -\infty$

Robot model

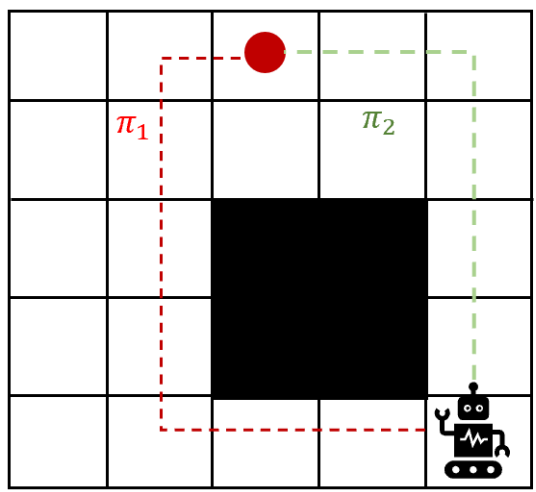

$C_R(\pi_{exp}) = -8$
$C_R(\pi_{opt}) = -6$

**2) The maps and costs in task 2 are**

$\pi_1 = \pi_{exp}$
$\pi_2 = \pi_{opt}$

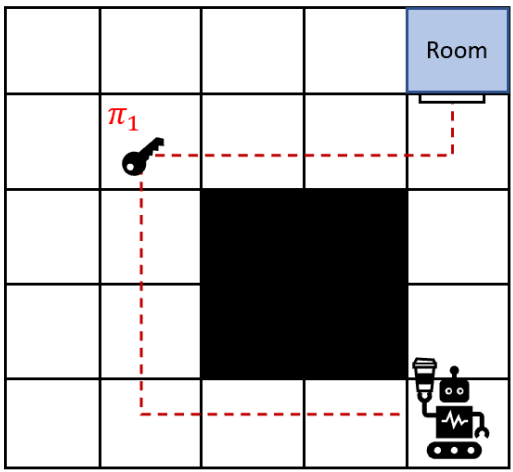
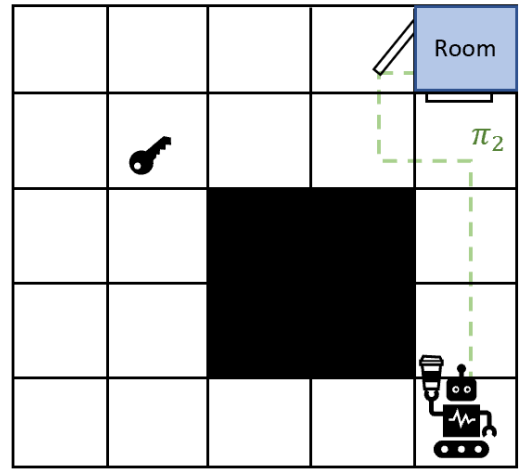

$C_H(\pi_{exp}) = -12$
$C_H(\pi_{opt}) = -\infty$

$C_R(\pi_{exp}) = -12$
$C_R(\pi_{opt}) = -6$

**3) The maps and costs in task 3 are**

$\pi_1 = \pi_{exp}$
$\pi_2 = \pi_{opt}$

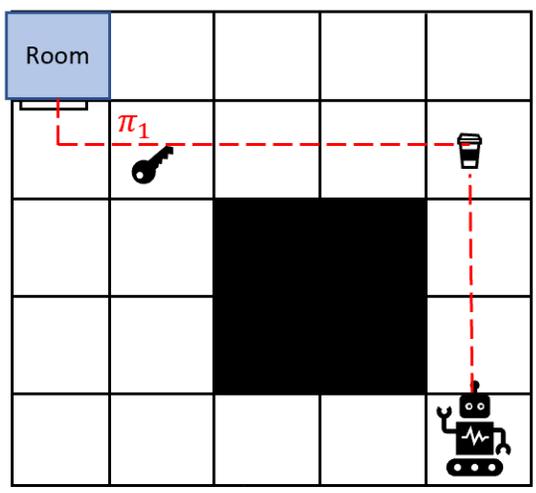 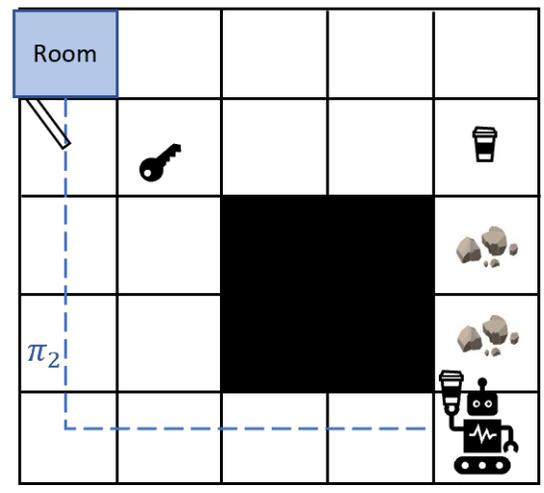

$C_H(\pi_{exp}) = -11$
$C_H(\pi_{opt}) = -\infty$

$C_R(\pi_{exp}) = -17$
$C_R(\pi_{opt}) = -8$

**4) The maps and costs in task 4 are**

$\pi_1 = \pi_{exp}$
$\pi_2 = \pi_{opt}$

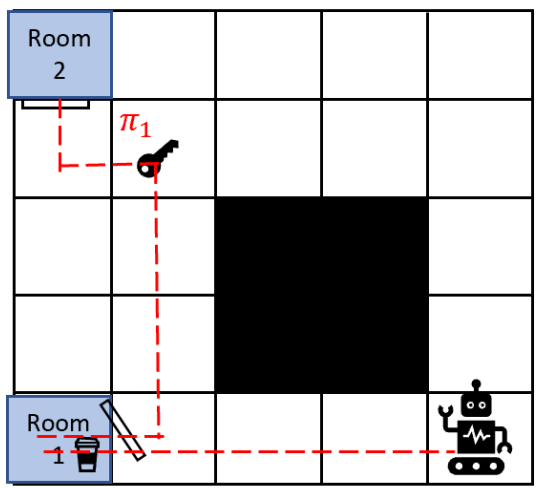
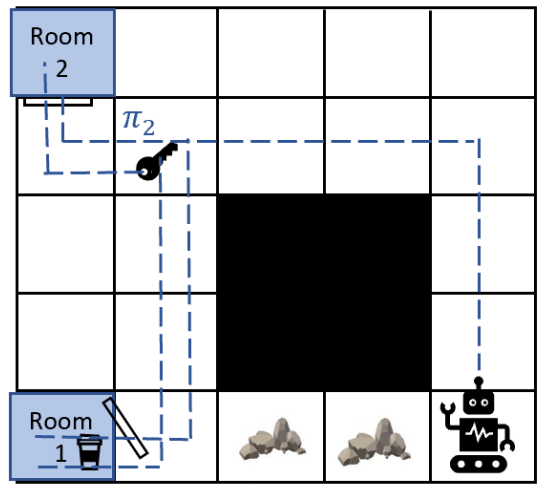

$C_H(\pi_{exp}) = -13$
$C_H(\pi_{opt}) = -19$

$C_R(\pi_{exp}) = -500$
$C_R(\pi_{opt}) = -19$

## Appendix C – Mars Rover Domain

In Rover domain, we mentioned that the Martian's model of the rover has an additional preconditions which are (empty ?s) for actions sample_soil, sample_rock, and sample_metal, and extra preconditions for the action take_image.

Now for each problem, the rover is expected to communicate soil, rock, metal, and images from a set of waypoints. Given the additional preconditions in the Martian model, the expected plan in the Martian model would be longer than what is required for the rover. For example, in the first problem, the rover goal consists of communicate_metal_data waypoint0 and communicate_metal_data waypoint3. For the first problem, the explicable and optimal plans are as follows

$\pi_{exp}^1 =$

```
(sample_metal rover0 rover0store waypoint3)
(communicate_metal_data rover0 general waypoint3 waypoint3 waypoint0)
(navigate rover0 waypoint3 waypoint0)
(drop rover0 rover0store)
(sample_metal rover0 rover0store waypoint0)
(navigate rover0 waypoint0 waypoint3)
(communicate_metal_data rover0 general waypoint0 waypoint3 waypoint0)
```

$\pi_{opt}^1 =$

```
(sample_metal rover0 rover0store waypoint3)
(communicate_metal_data rover0 general waypoint3 waypoint3 waypoint0)
(navigate rover0 waypoint3 waypoint0)
(sample_metal rover0 rover0store waypoint0)
(navigate rover0 waypoint0 waypoint3)
(communicate_metal_data rover0 general waypoint0 waypoint3 waypoint0)
```

## Appendix D – Human Study Details

In the human experiment setup, the users first should read through the instruction with the details about the study.
**The instruction is as follows.**

# Instructions

## Setup

Let's consider a scenario where you have a job supervising a robot. Here you are responsible for making sure the robot is performing their assigned tasks and is achieving their goals. In this study, we will turn this scenario into a game where you will have 10 attempts at supervising the robot. In each attempt, based on your choices you will be assigned some points, or you may lose some. In the end, depending on your score you will be awarded some bonus cash (this is in-addition to the base pay of $10). That is, you will get 1 cent for every 10 point you receive (with a maximum of $2).

As mentioned earlier, your primary responsibility is to ensure the robot completes its task. So, after each attempt, you will get 100 points if the robot does the task right and achieves the assigned goal.

At the beginning of each attempt, you can either choose to monitor the robot to make sure its does its job or you can choose to perform another task to make extra points.

In this case the extra task would be labeling images. For each round of labeling images, you will receive 100 points (in addition to the points you receive from the robot doing its tasks). But note that if you choose to label images you can't monitor the robot. And if the robot does something wrong and does not achieve the goal you will lose points (-200 points), though you still get to keep the points from your labeling task.

If you choose to monitor the robot and you see the robot doing something invalid or wrong, you can stop the robot with the stop button. Though in this scenario you will only receive 50 points. But if you choose to monitor the robot and let the robot finish its plan (i.e. not press the stop button at any point), in this case if the robot couldn't achieve the goal (either because at the end of the plan the robot doesn't get to the goal) then you will again lose points (-200 points).

## Costs

| # | Description | Points |
|---|---|---|
| 1 | Robot achieves the goal | +100 |
| 2 | Image Labelling | +100 |
| 3 | Robot fails to achieve the goal | -200 |
| 3 | Stopping the robot because doing wrong while monitoring | +50 |

## Monitoring

You will see a map like the one shown below.
Each map would have specific goal that the robot may need to achieve, which will be specified to you.
Over the 10 attempts, the robot may need to perform the same task over and over or may need to perform other task.

**For each task the robot should follow the shortest plan to achieve its goals.**

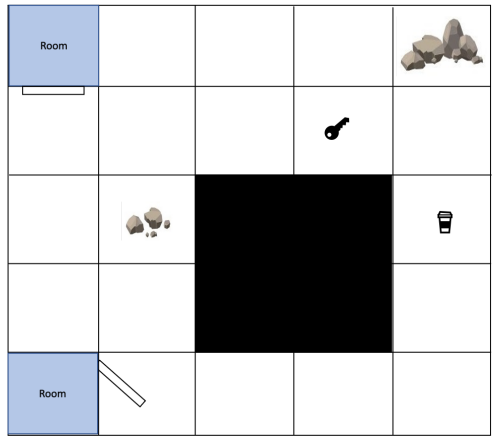

## Image Labelling

You will be shown different images of animals and objects and you will be asked to identify the class of the item/animal shown in the image from a set of option. Note, here you will not lose any point if you choose the wrong category. When you complete labeling you will be informed about whether the robot has successfully completed its task for that attempt or not.

At the end of each attempt, you will be asked to answer a short questionnaire. Before starting the attempts, we recommend that you go over the sample map to familiarize with it.

I Understand, Continue    Go to Top

After instruction page, a sample map is given to the participants to become familiar with the setup if they want. They can go through the sample map as many times as they need
**The sample map is**

# Sample Monitor Page

| Round # | Total score |
|---------|-------------|
| 1 | 0 |

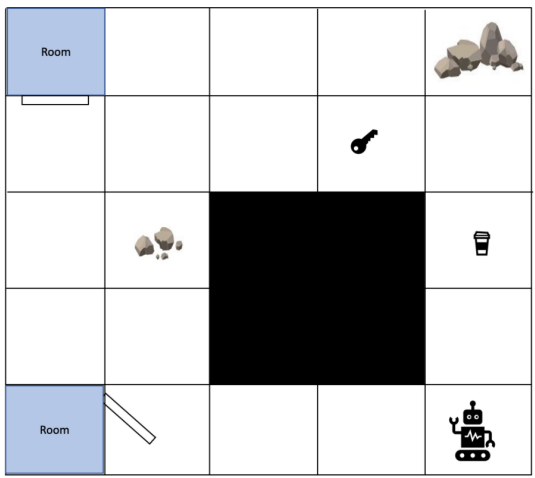

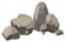 Rubbles that robot **Cannot** pass through them

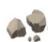 Rubbles that robot **Can** pass through them

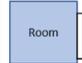 A room that the door is locked and need key to unlock

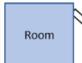 A room that the door is open and doesn't need key to go in

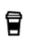 Coffee cups

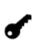 Key for locked doors

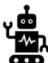 The Robot

- Please note that the Robot should take shortest path
- There might be a possibility that the robot might have a different understanding of the map(sensing or believing something you don't, but it's not necessarily true)
- Note that the robot should first take the key if the door is locked

After the sample map, they can start the survey, where it start with the question whether they want to monitor the robot or label the images.
The screenshot of the page is given.

# Make A Choice

Round # : 1

Do you want to monitor the agent or label the image?

Total score: 0

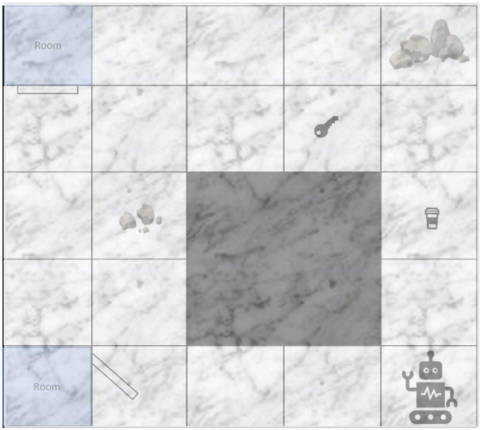

Monitor

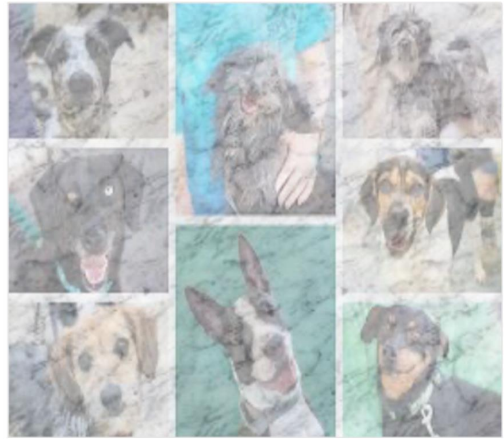

Label Images

If the user chooses monitor, they will go to the page with robot has a task to do. The users can stop the robot at any step or click on the "next" button to see next steps the robot executes till the end. An screenshot of the page is as follows.

# Montior Page

| Round # | The robot task is to reach to the red point on the map | Total score |
|---|---|---|
| 1 | | 0 |

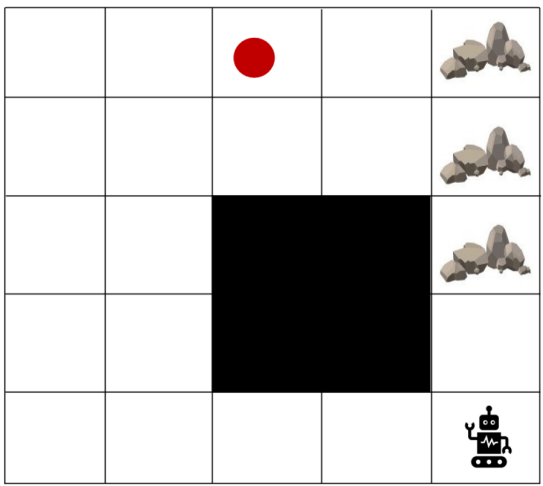

| | |
|---|---|
| 🪨 | Rubbles that robot **Cannot** pass through them |
| 🪨 | Rubbles that robot **Can** pass through them |
| Room | A room that the door is locked and need key to unlock |
| Room | A room that the door is open and doesn't need key to go in |
| ☕ | Coffee cups |
| 🔑 | Key for locked doors |
| 🤖 | The Robot |

- Please note that the Robot should take shortest path
- There might be a possibility that the robot might have a different understanding of the map(sensing or believing something you don't, but it's not necessarily true)
- Note that the robot should first take the key if the door is locked

[Next] [Stop]

But if the user chooses label, they will go to a page with some image that they should label them. The screenshot of the page is.

# Label Images

Round #     If the robot succeeds in its task, you will earn an extra 100 points for labelling the images.     Total score

1                                                                          0

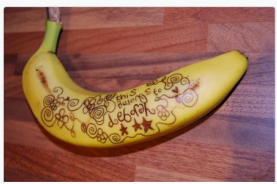

- ⦿ soccer
- ○ dog
- ○ truck
- ○ banana

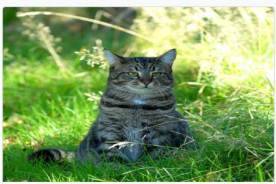

- ⦿ banana
- ○ cat
- ○ dog
- ○ truck

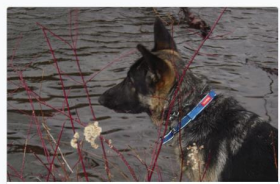

- ⦿ dog
- ○ plane
- ○ soccer
- ○ truck

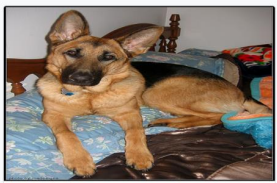

- ⦿ banana
- ○ bus
- ○ soccer
- ○ dog

Done

At the end of monitoring or labeling the users will be informed about whether the robot achieved its goal and the score they get in that round. At the end of each round a Muir Questionnaire is given to them to evaluate their trust. The users are given a Likert scale from 0 to 10 for each factor. Then the trust is measured by the average of these factors divided by 10. The Questionnaire is

# Trust Evaluation

Round #     Please answer the following questions for this round.     Total score

1                                                                                       200

To what extent can the robot's behavior be predicted from moment to moment?

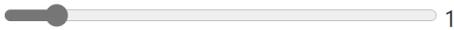 1

To what extent can you count on the robot to do its job?

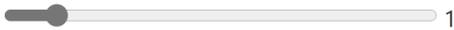 1

What degree of faith do you have that the robot will be able to cope with similar situations in the future?

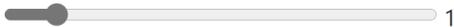 1

Overall, how much do you trust the robot?

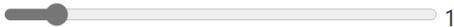 1

[Continue]

Given their trust value, a new or the same task is given to the robot in the next round. There are 10 rounds which has the same structure as described.



\end{document}